\documentclass[10pt,twocolumn,letterpaper]{article}

\usepackage{cvpr}              

\definecolor{cvprblue}{rgb}{0.21,0.49,0.74}
\usepackage[pagebackref,breaklinks,colorlinks,allcolors=cvprblue]{hyperref}

\def\paperID{9324} 
\def\confName{CVPR}
\def\confYear{2026}

\newcommand\myvspace{\vspace{0mm}}
\newcommand{\ours}{OnlinePG}

\colorlet{colorLow}{darkgray!50}    
\newcommand{\lo}[1]{{\textcolor{colorLow}{#1}}}      
\newcommand{\fs}[1]{{\textcolor{red}{\textbf{#1}}}}   
\newcommand{\nd}[1]{{\textcolor{green}{#1}}}      
\newcommand{\rd}[1]{{\textcolor{blue}{#1}}}      

\title{OnlinePG: Online Open-Vocabulary Panoptic Mapping with 3D Gaussian Splatting}

\usepackage{graphicx}
\usepackage{amsmath}
\usepackage{amssymb}
\usepackage{booktabs}
\usepackage{pifont}
\usepackage{multirow}
\usepackage{enumitem}
\usepackage[percent]{overpic}
\usepackage{verbatim}
\usepackage{bbm}
\usepackage{footnote}
\usepackage{tabularx}
\usepackage{arydshln} 
\usepackage{makecell}
\usepackage{subcaption} 

\author{Hongjia Zhai$^{1,2}$ \quad Qi Zhang$^{2}$ \quad Xiaokun Pan$^{1}$ \quad Xiyu Zhang$^{1}$ \quad Yitong Dong$^{1,2}$ \\ Huaqi Zhang$^{2}$ \quad Dan Xu$^{3}$ \quad Guofeng Zhang$^{1}$\footnotemark[2] \\
$^{1}$State Key Lab of CAD \& CG, Zhejiang University \quad $^{2}$VIVO BlueImage Lab \quad $^{3}$HKUST
}

\begin{document}
\twocolumn[{%
    \renewcommand\twocolumn[1][]{#1}%
    \vspace{-1.6em}
    \maketitle
    \centering
    \vspace{-0.5cm}
\includegraphics[width=0.99\linewidth]{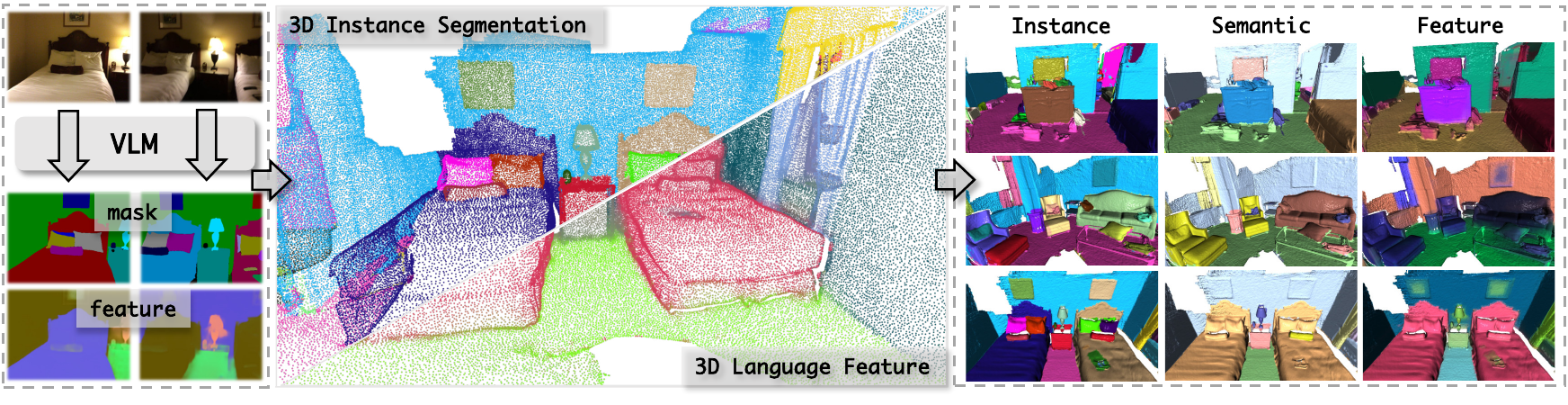}
\vspace{-3mm}
\captionof{figure}{\textbf{Illustration of \ours}, which integrates geometric reconstruction and open-vocabulary panoptic perception built upon 3D Gaussian Splatting. Given the posed video stream and 2D noise priors (mask and feature) generated by vision-language models (VLMs), {\ours} can reconstruct a consistent 3D language feature field and perform online open-vocabulary panoptic mapping.}
\label{fig:teaser}    
\vspace{4mm}
}]

\renewcommand{\thefootnote}{\fnsymbol{footnote}}
\footnotetext[2]{Corresponding author.}


\begin{abstract}
Open-vocabulary scene understanding with online panoptic mapping is essential for embodied applications to perceive and interact with environments. However, existing methods are predominantly offline or lack instance-level understanding, limiting their applicability to real-world robotic tasks. In this paper, we propose OnlinePG, a novel and effective system that integrates geometric reconstruction and open-vocabulary perception using 3D Gaussian Splatting in an online setting. Technically, to achieve online panoptic mapping, we employ an efficient local-to-global paradigm with a sliding window. To build local consistency map, we construct a 3D segment clustering graph that jointly leverages geometric and semantic cues, fusing inconsistent segments within sliding window into complete instances. Subsequently, to update the global map, we construct explicit grids with spatial attributes for the local 3D Gaussian map and fuse them into the global map via robust bidirectional bipartite 3D Gaussian instance matching. Finally, we utilize the fused VLM features inside the 3D spatial attribute grids to achieve open-vocabulary scene understanding. Extensive experiments on widely used datasets demonstrate that our method achieves better performance among online approaches, while maintaining real-time efficiency.
\end{abstract}

\section{Introduction}
\label{sec:intro}

Open-vocabulary 3D scene understanding is fundamental for embodied tasks, enabling robots to perceive, reason about, and interact with complex environments using natural language and instruction~\cite{huang23vlmaps,Peng2023OpenScene,shi2024_gs_language_embed,qin2024langsplat}.
While significant progress has been made in offline settings, real-world robotic applications demand online perception and mapping capabilities~\cite{mousavian20196,zhang20233d,monogs} that can process streaming data and support decision-making in real time.

To overcome the limited capability of annotated 3D data, recent works~\cite{wu2024panorecon,narita2019panopticfusion,kundu2022panopticneuralfiled,martins2024ovo,wang2025open} leverage various 2D vision-language models (VLMs)~\cite{clip,Lseg,dino,ravi2024sam2,oquab2023dinov2,jose2024dinov2meetstextunified} to transfer open-vocabulary knowledge into 3D space.
Building on differentiable rendering techniques such as Neural Radiance Fields (NeRF)~\cite{mildenhall:2020:nerf} and 3D Gaussian Splatting (3DGS)~\cite{kerbl3Dgaussians}, several methods~\cite{zhai_cvpr25_panogs,li2025instancegaussian,wu2024opengaussian,qin2024langsplat,shi2024_gs_language_embed,dong2026one} have achieved impressive results in fine-grained open-vocabulary scene understanding.
However, these approaches are predominantly offline and lack support for online instance-level panoptic perception, hindering their applications in embodied tasks.

Despite previous approaches that combine VLMs with 3DGS having yielded satisfactory performance, two critical limitations remain:
\textit{1) offline reconstruction and perception settings.}
Most existing reconstruction and understanding methods~\cite{shi2024_gs_language_embed,zhai_cvpr25_panogs,wu2024opengaussian,qin2024langsplat} require pre-collected data. 
This offline setting prevents robots from simultaneously exploring and understanding unknown environments in real time.
Online mapping and perception are indispensable for robots.
\textit{2) missing or inaccurate 3D instance-level understanding.}
Current online open-vocabulary scene understanding approaches~\cite{o2v-mapping_eccv24,yamazaki2024open_fusion} cannot distinguish individual 3D instances based on text queries, while offline instance-aware approaches~\cite{li2025instancegaussian,wu2024opengaussian,zhai_cvpr25_panogs,siddiqui2023panoptic-lifting} rely on time-consuming clustering or multi-view guidance that is unsuitable for online systems.
However, equipping the online mapping system with capabilities of open-vocabulary panoptic understanding is highly meaningful and needed for embodied applications.
Addressing these challenges is crucial for enabling real-time, open-vocabulary panoptic mapping and understanding in embodied applications.

To this end, we present {\ours}, an efficient online open-vocabulary panoptic mapping system based on 3D Gaussian Splatting that integrates geometric reconstruction with semantic understanding.
Our key insight is to employ an efficient local-to-global paradigm: we first build local consistent maps within a sliding window to mitigate noisy 2D VLM priors, then incrementally fuse them into a global representation.
Specifically, given noisy 2D masks and feature maps from VLMs, we initialize 3D Gaussian segments from keyframes and construct a multi-cue clustering graph that leverages geometric overlap, semantic similarity, and view consensus to merge inconsistent segments into complete instances.
We then voxelize the local map to build explicit spatial attribute grids for storing semantic features and instance labels, which are robustly fused into the global map via our bidirectional bipartite matching scheme.
This design enables accurate instance-level open-vocabulary understanding while maintaining real-time performance.

Overall, the technical contributions of our approach are summarized as follows:
\begin{itemize}
    \item We propose an online open-vocabulary panoptic mapping framework that unifies geometric reconstruction and semantic understanding in a local-to-global paradigm.
    \item We develop a multi-cue segment clustering algorithm that synergistically leverages geometric, semantic, and view consensus information to obtain local consistent 3D instances from noisy 2D VLM priors.
    \item We design a bidirectional bipartite matching scheme with explicit spatial attributes for robust incremental fusion of local 3D instances into the global map.
    \item Extensive experiments demonstrating our superior performance among online methods, with significant improvements in both semantic and panoptic metrics.
\end{itemize}

\section{Related Work}
\label{sec:related_work}

\noindent\textbf{Panoptic Segmentation.}
2D panoptic segmentation task was first proposed by Kirillov~\etal~\cite{kirillov2019panoptic}, which aims to understand objects with different semantics and instances within images.
Follow-up works~\cite{cheng2020panoptic,cheng2021mask2former,porzi2019seamless} have attempted to advance the CNN model's ability to reason and understand images from a human perspective.
To explore spatial semantics and understand 3D scenes, some works~\cite{narita2019panopticfusion,zhai_cvpr25_panogs,siddiqui2023panoptic-lifting,fu2022panopticnerf,kundu2022panopticneuralfiled,zhu2025pcf-lifting} integrate and lift 2D panoptic predictions of 2D models~\cite{cheng2021mask2former,cropformer} into 3D space.
Inspired by~\cite{mccormac2017semanticfusion}, PanopticFusion~\cite{narita2019panopticfusion} fuses the 2D panoptic information into 3D volume and performs map regularization.
In addition, some works~\cite{kundu2022panopticneuralfiled,zhai_cvpr25_panogs,li2025instancegaussian} have exploited recent optimization and rendering capabilities of differentiable representations~\cite{mildenhall:2020:nerf,kerbl3Dgaussians} to understand 3D scenes.
However, limited by the offline or closed-set setting, these works cannot simultaneously perform online 3D panoptic reconstruction or open vocabulary 3D scene understanding.

\noindent\textbf{Online 3D Scene Mapping and Perception.}
For real-world applications of embodied intelligence~\cite{mousavian20196,zhang20233d,nis-slam,huang23vlmaps}, intelligent robots typically perceive and understand the environment with an online data stream.
Previous online 3D scene mapping and perception methods~\cite{xu2024embodiedsam,point-slam,martins2024ovo,tang2025onlineanyseg,yamazaki2024open_fusion,narita2019panopticfusion} were generally implemented by projecting 2D semantic predictions back into 3D scene maps.
Accurately reconstructing and understanding the surrounding 3D scene from RGB-D video streams is the visual foundation for these robotic tasks.
Benefiting from NeRF/3DGS-based reconstruction~\cite{zhai2025splatloc,Vox-Surf,zhang2025atlasgs,zhai2025neuraloc}, recent works~\cite{co-slam,nis-slam,vox-fusion,vox-fusion++,o2v-mapping_eccv24,pan2025egg,OGScene3D} have incorporated NeRF and 3DGS for fine-grained, better online mapping and perception performance.
However, these online approaches cannot achieve instance-aware panoptic mapping and perception for open-vocabulary scene understanding.

\begin{figure*}[ht!]
\centering
\includegraphics[width=\linewidth]{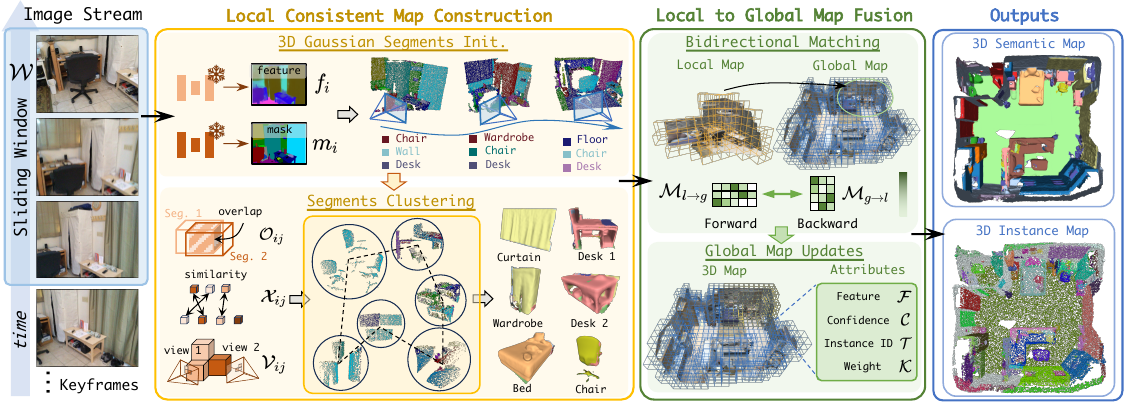} 
\caption{\textbf{Overview of Our Approach}. 
Our system performs online open-vocabulary panoptic mapping from RGB-D streams using a local-to-global paradigm. 
\textbf{(a)} Maintaining a sliding window $\mathcal{W}$ to build locally consistent instance and fuse them globally. 
\textbf{(b)} Using noisy 2D masks and features from~\cite{cropformer,Lseg} to initialize 3D Gaussian segments, which are clustered via multi-cues to form consistent 3D instances. 
\textbf{(c)} Bidirectional bipartite matching incrementally fuses local instances and spatial attributes into the global map.}
\label{fig:pipline}
\end{figure*}
\noindent\textbf{Open Vocabulary 3D Scene Understanding.}
3D scene understanding plays an important role in embodied intelligence.
However, limited 3D annotated datasets~\cite{dai:2017:scannet,julian:2019:replica} have hindered the development of open-vocabulary 3D scene understanding, with most works~\cite{vu2022softgroup,mask3d,xu2024embodiedsam} only achieving good results in closed-vocabulary settings.
With the rapid advancement of 2D VLM~\cite{clip,dino,Lseg}, some efforts~\cite{Peng2023OpenScene,zhai_cvpr25_panogs} use 3D consistent feature fields to model the scene property with the lifted 2D VLM features as supervision information.
LERF~\cite{lerf} and N3F~\cite{n3f} are the first batch of works that combine NeRF and VLMs for open vocabulary 3D scene understanding, learning consistent feature fields in 3D space through volume rendering.
And the follow-up works~\cite{guo2024semantic_gaussian,shi2024_gs_language_embed,zuo2024fmgs,zhou2024feature_3dgs,chacko2025lifting_by_gaussian} replace NeRF with 3DGS for efficient rendering and explicit 3D scene representation.
Besides, to distinguish the instance information of 3D Gaussian primitives, the recent PanoGS~\cite{zhai_cvpr25_panogs} and InstanceGaussian~\cite{li2025instancegaussian} use 2D instance information of SAM~\cite{sam} to guide the 3D Gaussian clustering.
Those offline approaches adopt contrastive feature learning~\cite{gaussian_grouping,wu2024opengaussian,guo2024semantic_gaussian} to group 3D Gaussian primitives into complete instances.
Due to the inherently slow convergence of this mechanism, adapting it to online mapping systems will result in inaccurate 3D instance results.
Therefore, their offline setting hinders scene understanding applications for online embodied tasks.

\section{Method}
\label{sec:method}

As shown in~\cref{fig:pipline}, given the posed RGB-D stream, we can perform 3D online open-vocabulary panoptic mapping, which formulates the pipeline as an efficient local-to-global paradigm.
To mitigate the inconsistencies of 2D segmentation results, we propose an effective segment clustering algorithm that synergistically leverages geometric and semantic cues to obtain consistent 3D Gaussian instances.
Then we construct a grid-based spatial attribute that stores semantic information of the reconstructed scene (\cref{subsec:local}).
Finally, we design a bidirectional bipartite matching scheme to incrementally fuse local 3D Gaussian instances and spatial attribute grids into global map
(\cref{subsec:global}).

\subsection{Scene Representation}
\label{subsec:scene_representation}

\noindent\textbf{3D Gaussian Splatting.}
Benefiting from the efficiency of 3DGS~\cite{kerbl3Dgaussians}, we represent the 3D scene with 3D Gaussians for online open vocabulary panoptic mapping.
We incrementally reconstruct the 3D Gaussian primitives from the RGB-D stream:
\begin{equation}
    \mathcal{G}_i:=\left\{\boldsymbol{\mu}_{i}, \boldsymbol{\Sigma}_{i}, \sigma_{i}, \boldsymbol{c}_{i}\right\},
\end{equation}
where each Gaussian primitive $\mathcal{G}_i$ contains its position $\boldsymbol{\mu}_{i} \in \mathbb{R}^3$, covariance matrix $\boldsymbol{\Sigma}_{i} \in \mathbb{R}^{3 \times 3}$, opacity $\sigma_i \in \mathbb{R}$, and color $\boldsymbol{c}_{i} \in \mathbb{R}^{3}$.

The 3D scene properties can be rasterized to the 2D image plane via fast differentiable rasterization~\cite{max1995optical} for differentiable optimization.
Following previous works~\cite{monogs,kerbl3Dgaussians}, we adopt the  L1 loss terms for appearance and geometry optimization:
\begin{equation}
\mathcal{L} = \alpha \cdot \mathcal{L}_{c} + (1-\alpha) \cdot \mathcal{L}_{d},
\label{eq:geo_color}
\end{equation}
where $\alpha$ is the weight for the optimization components.

\noindent\textbf{Grid-based Spatial Attributes.}
In addition to these anisotropic attributes bound to Gaussian primitives, we require additional representations to model open-vocabulary panoptic information.
Therefore, for efficient computation and lightweight storage, we perform voxelization on the reconstructed regions and adopt explicit sparse grids to record spatial geometric and semantic attributes.
The spatial attributes are represented by a set of sparse voxel grids, where each occupied voxel contains $\{\mathcal{F}, \mathcal{C},\mathcal{T},\mathcal{K}\}$: $\mathcal{F} \in \mathbb{R}^{D_f}$ stores language features extracted from VLMs, $\mathcal{C} \in \mathbb{R}$ is the confidence of the feature, $\mathcal{T} \in \mathbb{R}$ is the panoptic instance label, and $\mathcal{K}  \in \mathbb{R}$ is the instance weight.
These grids are detailed in the following sections.

\subsection{Local Consistent Map Construction}
\label{subsec:local}
To obtain accurate 3D Gaussian instances, existing methods adopt contrastive feature learning~\cite{li2025instancegaussian,wu2024opengaussian,gaussian_grouping} guided by inconsistent 2D masks.
However, this paradigm is unsuitable for online reconstruction due to slow convergence and time-consuming post-processing.
Since 2D segmentation priors are inherently noisy (\eg, multi-view inconsistency, over-segmentation), directly lifting and fusing 2D keyframes into the 3D map~\cite{narita2019panopticfusion,mccormac2017semanticfusion} is both inaccurate and inefficient.
Therefore, we maintain a sliding window over the input stream and perform 3D segment clustering to obtain consistent instances while mitigating noise from 2D priors.

\myvspace\noindent\textbf{3D Gaussian Segments Initialization.}
For $i$-th keyframe inside the sliding window $\mathcal{W}$, we use LSeg~\cite{Lseg} and EntitySeg~\cite{cropformer} to extract its 2D feature map $f_i\in \mathbb{R}^{H \times W \times D_f}$ and instance mask $m_i \in \mathbb{R}^{H \times W}$. 
We then lift the keyframe into 3D space using its depth value to initialize 3D Gaussian primitives.
The Gaussian primitives are grouped into multiple 3D segments according to their corresponding 2D mask IDs: primitives sharing the same mask ID in $m_i$ are assigned to the same segment.
For simplicity, we refer to a group of Gaussian primitives with the same instance ID as a 3D segment $\mathcal{S}_i$.
In this way, we obtain multiple 3D segments initialized from keyframes inside the sliding window.
\begin{equation}
    \mathcal{S} := \left\{\mathcal{S}_1, \cdots, \mathcal{S}_n, | \mathcal{S}_i:=\left\{ \mathcal{G}_j \right\}_{j=1}  \right\},
\end{equation}
where $|m_i|$ is the 2D instance number of $I_i$, and $n = \sum_{i \in \mathcal{W}} |m_i|$ is the 3D segment number insides the window.

Due to the inconsistency of 2D segmentation $m_i$, the resulting 3D segments $\mathcal{S}$ may be over-segmented or under-segmented.
To obtain consistent instances from $\mathcal{W}$ for subsequent reconstruction and rendering, we construct a clustering graph for the inconsistent $\mathcal{S}$ and cluster them based on their mutual relationships to obtain the consistent 3D Gaussian instances $\mathcal{I}$.

\myvspace\noindent\textbf{Segments Clustering Graph with Multi-cues.}
We formulate the clustering graph as $(\{\mathcal{S}_i\}, \{\mathcal{E}_{ij}\})$, where 3D segments are defined as the vertices $\mathcal{S}_i$, and the edges $\mathcal{E}_{ij}$ represent the affinity between different segments, reflecting the probability that segments belong to the same instance.
To capture the detailed affinity, we take the geometry, semantic, and view consensus cues into consideration.

\textbf{(1) Geometry cue}:
Since each 3D segment contains multiple 3D Gaussian primitives with varying densities, traversing all primitives for overlap computation is time-consuming.
To efficiently calculate the overlap between different Gaussian segments, we uniformly divide the space into axis-aligned voxels of specified size and count the overlapped voxels instead:
\begin{equation}
\mathcal{O}(\mathcal{S}_i, \mathcal{S}_j) = \frac{1}{2} \cdot \left(\frac{|\mathcal{S}_i \cap \mathcal{S}_j |}{\mathrm{Cont.}(\mathcal{S}_i, \mathcal{S}_j)} + \frac{|\mathcal{S}_i \cap \mathcal{S}_j |}{\mathrm{Cont.}(\mathcal{S}_j, \mathcal{S}_i)}\right),
\end{equation}
where $\mathrm{Cont.}(\mathcal{S}_i, \mathcal{S}_j)$ denotes the ratio of visible voxels in $\mathcal{S}_j$ that are contained in $\mathcal{S}_i$ when projecting $\mathcal{S}_j$ back to the viewpoint of $\mathcal{S}_i$.

\textbf{(2) Semantic cue}:
To obtain the language feature for segment $\mathcal{S}_i$, we pool the language feature map according to its 2D mask: $z_i = \Phi(\{f(u,v): m(u,v)=i\})$, where $\Phi(\cdot)$ denotes average pooling.
The semantic cue is then computed as the cosine similarity between language features: $\mathcal{X}(\mathcal{S}_i, \mathcal{S}_j) = z_i \cdot z_j /(||z_i|| \cdot ||z_j||)$.

\textbf{(3) View consensus cue}: We use the view consensus rate from~\cite{yan2024maskclustering} as additional criteria to judge whether two segments belong to the same 3D instance.
\begin{equation}
    \mathcal{V}(\mathcal{S}_i, \mathcal{S}_j) = \frac{N_{\mathrm{supp}}(\mathcal{S}_i, \mathcal{S}_j)}{N_{\mathrm{vis}}(\mathcal{S}_i, \mathcal{S}_j)},
\end{equation}
where $N_{\mathrm{supp}}$ and $N_{\mathrm{vis}}$ represent the number of keyframes where both segments are contained and visible.

\myvspace\noindent\textbf{Clustering for Local Consistent Map.}
Once the clustering graph $(\{\mathcal{S}_i\}, \{\mathcal{E}_{ij}\})$ is constructed, we perform graph-based clustering to merge over-segmented and under-segmented results from the sliding window into consistent 3D instances.
To determine whether two segments should be merged, we adopt the following criteria:
\begin{equation}
    \Delta_{ij} = \left(( \mathcal{O}_{ij} + \mathcal{X}_{ij} ) > \lambda_1 \right) \cup \left(\mathcal{V}_{ij} > \lambda_2 \right),
    \label{eq:delta}
\end{equation}
where $\lambda _{1}$ and $\lambda _{2}$ are the threshold parameters controlling the merging decision.

We iterate through the graph using~\cref{eq:delta} to identify connected components, with each component representing a consistent instance.
The final consistent 3D Gaussian instances $\mathcal{I}$ are obtained via:
\begin{equation}
    \mathcal{I} = \mathrm{Cluster}(\{\mathcal{S}_i\}, \{\mathcal{E}_{ij}:=\{ \mathcal{O}_{ij}, \mathcal{X}_{ij}, \mathcal{V}_{ij} \} \}),
\end{equation}
where $\mathrm{Cluster}(\cdot)$ denotes the clustering operation based on connected components.

Through this multi-cue graph clustering algorithm, we obtain geometrically and semantically consistent 3D Gaussian instances $\mathcal{I}$ from the local sliding window.

\myvspace\noindent\textbf{Local Spatial Attribute Grids.}
After clustering the 3D Gaussian segments, we voxelize the 3D space to efficiently compute and update spatial attributes.
For each voxel $v$ occupied by instance $\mathcal{I}_i$, we assign the local instance label and weight grids:
\begin{equation}
    \mathcal{T}_{l}^t(v) = \mathrm{ID}_i, \quad \mathcal{K}_{l}^t(v) = N_i,
\end{equation}
where $t$ denotes the time index, $l$ denotes the local map, $\mathrm{ID}_i$ is the instance ID of $\mathcal{I}_i$, and $N_i$ is the number of observations of $\mathcal{I}_i$ during clustering, reflecting its confidence and frequency.

To enable open-vocabulary scene understanding, we aggregate multi-view language features from the sliding window $\mathcal{W}$ into local feature grid $\mathcal{F}_l^t$ and confidence grid $\mathcal{C}_l^t$:
\begin{equation}
    \mathcal{F}_{l}^t(v) = \frac{1}{\mathcal{C}_l^t(v)} \sum_{i \in \mathcal{W}} c_i(u,v) \cdot f_i(u,v),
\end{equation}
where $\mathcal{C}_l^t(v) = \sum_{i \in \mathcal{W}} c_i(u,v)$ and $c_i$ is the confidence map of feature $f_i$. 

\subsection{Local-to-Global Map Fusion}
\label{subsec:global}
After reconstructing the local 3D Gaussian instance map and spatial attribute grids from the sliding window $\mathcal{W}$, we match and register them into the global map.
To establish accurate matches between local and global maps, we leverage geometric and semantic information from the spatial attribute grids to construct correspondence score matrix.

\myvspace\noindent\textbf{Bidirectional Bipartite Matching.}
To establish one-to-one correspondences between local and global maps, we first construct a forward correspondence score matrix $\mathcal{M}_{l \to g} \in \mathbb{R}^{n_l \times n_g}$:
\begin{equation}
\mathcal{M}_{l \to g} = \frac{z_l \cdot z_g}{||z_l|| \cdot ||z_g||} + \frac{|\mathcal{I}_l \cap \mathcal{I}_g |}{\mathrm{Cont.}(\mathcal{I}_l, \mathcal{I}_g)},
\label{eq:match_l2g}
\end{equation}
where $n_l$ and $n_g$ are the numbers of instances in the local and global maps, and $z_l$ is the instance-level language feature of $\mathcal{I}_l$ queried from feature grid $\mathcal{F}_l$.

However, since the local map contains newly explored regions while the global map contains historical regions, the correspondence matrix is inherently asymmetric, particularly in geometric containment ratios.
To achieve robust matching, we construct a backward correspondence matrix $\mathcal{M}_{g \to l} \in \mathbb{R}^{n_g \times n_l}$ by replacing the second term in~\cref{eq:match_l2g} with $(|\mathcal{I}_g \cap \mathcal{I}_l |) / \mathrm{Cont.}(\mathcal{I}_g, \mathcal{I}_l)$.

With the above bidirectional matching matrix $\mathcal{M}_{l \to g}$ and $\mathcal{M}_{g \to l}$, we perform forward and backward bipartite graph matching via the Hungarian algorithm~\cite{kuhn1955hungarian} to yield the robust matching results:
\begin{equation}
    \mathcal{A} = \mathrm{Hung.}(\mathcal{M}_{l \to g}) \cap \mathrm{Hung.}(\mathcal{M}_{g \to l})^{T},
\end{equation}
where $\mathcal{A}$ is the final set of matched correspondences, $\mathrm{Hung.}(\cdot)$ is the Hungarian algorithm, and $(\cdot)^{T}$ denotes transpose to ensure bidirectional consistency.
This bidirectional strategy ensures reliable matching by requiring mutual agreement between forward and backward directions.

\myvspace\noindent\textbf{Global Map Update.}
For each voxel $v$ occupied by a clustered instance $\mathcal{I}$, we update the global feature grid $\mathcal{F}_g^t$ and confidence grid $\mathcal{C}_g^t$ using weighted averaging:
\begin{equation}
    \mathcal{F}_{g}^{t}(v) = \frac{\mathcal{C}_{l}^{t}(v) \cdot \mathcal{F}_{l}^{t}(v) + \mathcal{C}_{g}^{t-1}(v) \cdot \mathcal{F}_{g}^{t-1}(v)}{\mathcal{C}_{g}^{t}(v)},
\end{equation}
where $\mathcal{C}_{g}^{t}(v) = \mathcal{C}_{l}^{t}(v) + \mathcal{C}_{g}^{t-1}(v)$ is the updated confidence.

Since spatial attributes store discrete instance information in voxel form, we cannot perform continuous gradient updates.
Instead, we update the instance label $\mathcal{T}_g^t$ and weight $\mathcal{K}_g^t$ grids similar to~\cite{narita2019panopticfusion,mccormac2017semanticfusion}:
\begin{itemize}
\item \textbf{For matched instances:} 
We merge the matched instances into the global map by increasing the weight for each occupied voxel $v$:
\begin{equation}
    \mathcal{T}_{g}^{t}(v)=\mathcal{T}^{t-1}_{g}(v), \quad \mathcal{K}_{g}^{t}(v)=\mathcal{K}^{t-1}_{g}(v) + \mathcal{K}_{l}^{t}(v) .
\end{equation}
\item \textbf{For unmatched instances:} If $\mathcal{K}_{l}^{t}(v) \leq \mathcal{K}^{t-1}_{g}(v)$, we retain the global instance but decrease its weight:
\begin{equation}
    \mathcal{T}_{g}^{t}(v)=\mathcal{T}^{t-1}_{g}(v), \quad \mathcal{K}_{g}^{t}(v)=\mathcal{K}^{t-1}_{g}(v) - \mathcal{K}_{l}^{t}(v) .
\end{equation}
If $\mathcal{K}_{l}^{t}(v) > \mathcal{K}^{t-1}_{g}(v)$, indicating the local segmentation is more reliable, we replace the global instance:
\begin{equation}
    \mathcal{T}_{g}^{t}(v)=\mathcal{T}^{t}_{l}(v), \quad \mathcal{K}_{g}^{t}(v) = \mathcal{K}_{l}^{t}(v) - \mathcal{K}^{t-1}_{g}(v) .
\end{equation}
\end{itemize}

Through our bidirectional bipartite matching and robust fusion strategy, we can effectively integrate local 3D Gaussian instances and spatial attributes into the global map, enabling accurate and consistent online open-vocabulary panoptic mapping.

\section{Experiments}
\label{sec:exps}
In this section, we introduce our experimental settings and present open-vocabulary panoptic mapping results on two commonly used datasets. 
Additionally, we perform a detailed ablation study to validate the effect of each design in our system.


\begin{table*}[t]
\caption{\textbf{3D Semantic and Panoptic Segmentation Results on ScanNetV2 and Replica Datasets.} $*$ indicates the baseline results are taken from~\cite{zhai_cvpr25_panogs} which use the 3D instance obtained from supervised approach~\cite{vu2022softgroup} for panoptic comparison (their panoptic results are marked in \lo{gray}). We mark the top-3 segmentation results among all approaches in \fs{red}, \nd{green}, and \rd{blue}, respectively.}
\centering
\small
\setlength{\tabcolsep}{0.6em}
\begin{tabular}{lcc|cccc|cccc}
\toprule
    \multirow{2}{*}{Methods} & \multirow{2}{*}{Online} & \multirow{2}{*}{Pano.} & \multicolumn{4}{c|}{ScanNetV2~\cite{dai:2017:scannet}} & \multicolumn{4}{c}{Replica~\cite{julian:2019:replica}} \\
    & & & mIoU$\uparrow$ & mAcc.$\uparrow$ & PRQ (T)$\uparrow$ & PRQ (S)$\uparrow$ & mIoU$\uparrow$ & mAcc.$\uparrow$ & PRQ (T)$\uparrow$ & PRQ (S)$\uparrow$\\
    \midrule
    LangSplat~\cite{qin2024langsplat}$^*$ & \textcolor{red}{\ding{55}} & \textcolor{red}{\ding{55}} & 29.47 & 45.29 & \lo{22.57} & \lo{28.44} & 4.82 & 10.03 & \lo{8.29} & \lo{1.28} \\
    OpenGaussian~\cite{wu2024opengaussian}$^*$ & \textcolor{red}{\ding{55}} & \textcolor{red}{\ding{55}} & 24.89 & 37.35 & \lo{22.87} & \lo{19.71} & -- & -- & -- & -- \\
    OpenScene~\cite{Peng2023OpenScene}$^*$ & \textcolor{red}{\ding{55}} & \textcolor{red}{\ding{55}} & \rd{47.63} & \nd{69.74} & \lo{43.53} & \lo{40.43} & \nd{49.03} & \nd{62.89} & \lo{33.04} & \lo{11.84} \\
    InstanceGaussian~\cite{li2025instancegaussian} & \textcolor{red}{\ding{55}} & \textcolor{ForestGreen}{\ding{51}} & 34.14 & 54.95 & \fs{39.04} & \rd{27.41} & -- & -- & -- & -- \\
    PanoGS~\cite{zhai_cvpr25_panogs} & \textcolor{red}{\ding{55}} & \textcolor{ForestGreen}{\ding{51}} & \fs{50.72} & \fs{70.20} & 33.84 & \nd{36.22} & \fs{54.98} & \fs{67.35} & \fs{43.04} & \fs{30.60} \\
    \noalign{\vskip 1.5pt} \hdashline \noalign{\vskip 1.5pt}
    O2V-Mapping~\cite{o2v-mapping_eccv24} & \textcolor{ForestGreen}{\ding{51}} & \textcolor{red}{\ding{55}} & 33.74 & 55.52 & -- & -- & 24.35 & 33.62 & -- & --\\
    OnlineAnySeg~\cite{tang2025onlineanyseg} & \textcolor{ForestGreen}{\ding{51}} & \textcolor{ForestGreen}{\ding{51}} & 31.28 & 52.20 & \rd{35.98} & 26.27 & 37.48 & 45.64 & \rd{34.19} & \rd{9.13} \\
    Ours & \textcolor{ForestGreen}{\ding{51}} & \textcolor{ForestGreen}{\ding{51}} &  \nd{48.48} & \rd{66.01} & \nd{37.97} & \fs{41.81} & \rd{47.93} & \rd{54.94} & \nd{41.02} & \nd{12.83} \\
\bottomrule
\end{tabular}
\label{tab:main_results}
\end{table*}

\begin{figure*}[ht!]
\centering
 \includegraphics[width=\linewidth]{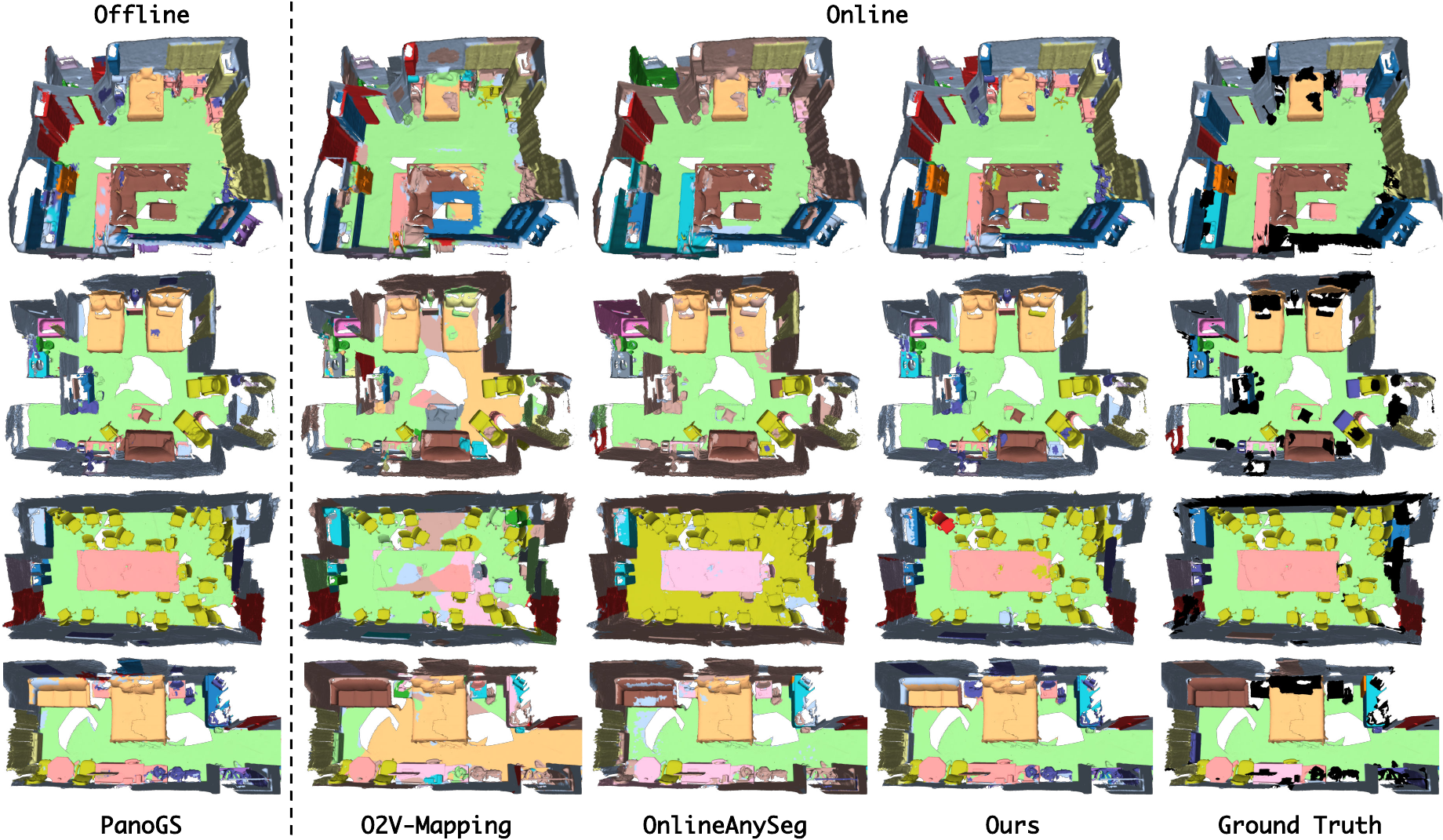} 
\vspace{-5mm}
\caption{\textbf{Qualitative 3D Semantic Segmentation Comparison of ScanNetV2 Dataset}. Our approach outperforms recent online approaches, O2V-Mapping~\cite{o2v-mapping_eccv24} and OnlineAnySeg~\cite{tang2025onlineanyseg}, by a large margin. Compared with the offline SOTA PanoGS~\cite{zhai_cvpr25_panogs}, our approach still exists a small performance gap.}
\label{fig:scannet_semantic_mesh}
\end{figure*}
\begin{figure*}[ht!]
\centering
\includegraphics[width=\linewidth]{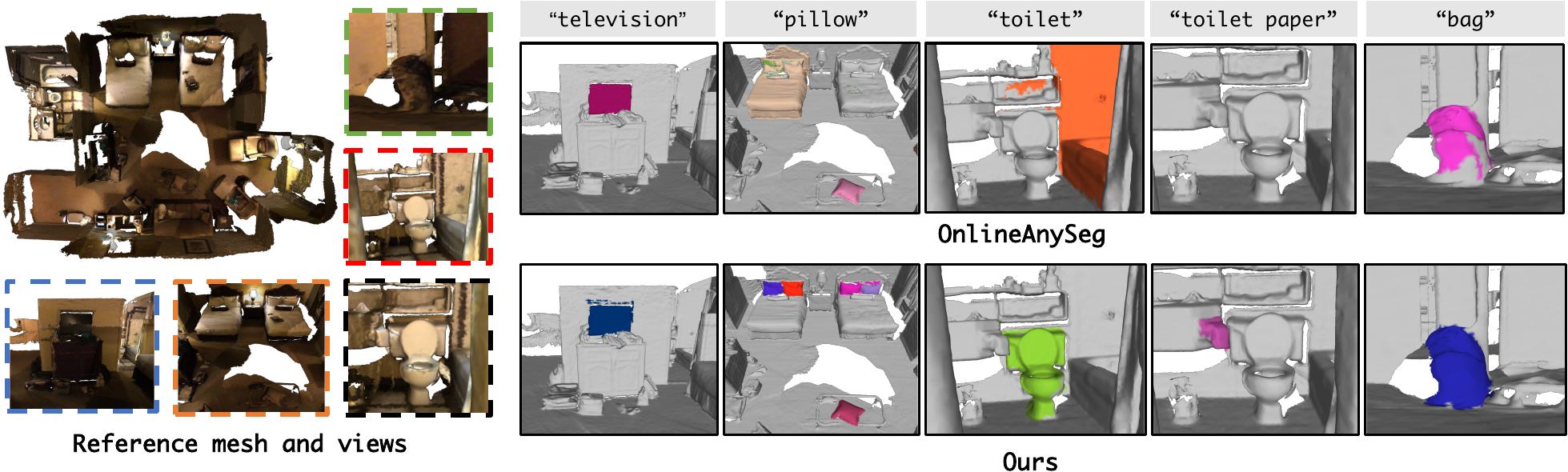} 
\caption{\textbf{Qualitative Results of Open-Vocabulary Query}. We use different colors to distinguish different instances found in the query. Compared to OnlineAnySeg~\cite{tang2025onlineanyseg}, our approach shows advanced text query capabilities for long-tail and multiple instances scenario.}
\label{fig:exp_open_query}
\end{figure*}

\subsection{Experimental Settings}
\label{subsec:exp_setting}
\myvspace\noindent\textbf{Datasets.}
Following the previous works~\cite{wu2024opengaussian,zhai_cvpr25_panogs,o2v-mapping_eccv24,Peng2023OpenScene}, we evaluate the quantitative and qualitative performance on a variety of scenes from two commonly used datasets, ScanNetV2~\cite{dai:2017:scannet} and Replica~\cite{julian:2019:replica}.
The two datasets both contain high-quality RGB-D sequences of various indoor scenes and 3D instance-level semantic annotations. 
Following~\cite{wu2024opengaussian,zhai_cvpr25_panogs}, we take the commonly-used 8 scenes $\{\texttt{room0-2,office0-4}\}$ for Replica dataset.
For the ScanNetV2 dataset, we use the same 10 selected sequences and settings for evaluation.

\myvspace\noindent\textbf{Evaluation Metrics.}
For open-vocabulary panoptic mapping evaluation, we adopt four widely-used metrics: 3D point-level mean Intersection over Union (mIoU), mean Accuracy (mAcc), and 3D Panoptic Reconstruction Quality at \textit{thing}-level (PRQ (T)) and \textit{stuff}-level (PRQ (S)).
Unlike offline baselines~\cite{qin2024langsplat,wu2024opengaussian,zhai_cvpr25_panogs} that use ground truth point clouds as input, our method reconstructs the scene from RGB-D streams.
Therefore, we project our reconstructed results to the ground truth point clouds for evaluation.

\myvspace\noindent\textbf{Implementation Details.}
Following~\cite{Peng2023OpenScene,zhai_cvpr25_panogs}, we adopt CLIP~\cite{clip} and LSeg~\cite{Lseg} as text and image visual-language feature extractors, with feature dimension $D_f=512$.
We use EntitySeg~\cite{cropformer} to extract 2D instance segmentation for each keyframe.
The hyperparameters are set as: $\alpha=0.9$, $\lambda_1=1.5$, $\lambda_2=0.8$.
We sample a keyframe every 20 frames and maintain a sliding window of size 12.
Segment clustering and local-to-global map fusion are performed every 7 keyframes.
The voxel size for spatial attributes is 3 cm.
For 3D Gaussian rendering optimization, when adding a new keyframe, we randomly select 5 historical keyframes and perform 20 optimization iterations using~\cref{eq:geo_color}.

\myvspace\noindent\textbf{Baselines.}
To compare our approach with recent open-vocabulary methods, we take five offline approaches: LangSplat~\cite{qin2024langsplat}, OpenGaussians~\cite{wu2024opengaussian}, OpenScene~\cite{Peng2023OpenScene}, InstanceGaussian~\cite{li2025instancegaussian}, PanoGS~\cite{zhai_cvpr25_panogs}, and two recent online approaches: O2V-Mapping~\cite{o2v-mapping_eccv24}, OnlineAnySeg~\cite{tang2025onlineanyseg} as baselines.
Since the baselines~\cite{qin2024langsplat,wu2024opengaussian,Peng2023OpenScene} marked with $*$ cannot obtain 3D panoptic results, we use the performance reported in~\cite{zhai_cvpr25_panogs}, which uses a supervised 3D instance segmentation approach~\cite{vu2022softgroup} (trained on ScanNetV2~\cite{dai:2017:scannet}) to obtain 3D instances for them.
So, we mark their PRQ performance in \lo{gray}.

\subsection{Main Experiments}
\label{subsec:main_exp}
We evaluate the 3D panoptic segmentation metrics of our approach and baseline on two commonly used ScanNetV2~\cite{dai:2017:scannet} and Replica~\cite{julian:2019:replica} datasets.
The averaged quantitative 3D semantic and panoptic segmentation results are shown in~\cref{tab:main_results}.
We mark the top-3 segmentation results among all approaches in \fs{red}, \nd{green}, and \rd{blue}, respectively.

\myvspace\noindent\textbf{3D Semantic Segmentation.}
As shown in~\cref{tab:main_results}, our method achieves the best 3D semantic segmentation results among online approaches on the mIoU and mAcc metrics of two datasets.
Compared to O2V-Mapping~\cite{o2v-mapping_eccv24} and OnlineAnySeg~\cite{tang2025onlineanyseg}, by maintaining and updating voxel-level spatial language feature grid $\mathcal{F}$, we can achieve more fine-grained 3D scene understanding performance, thereby avoiding language feature drift caused by fusing features from the noise instance parts.
Compared to the best current offline method, OnlineAnySeg~\cite{tang2025onlineanyseg}, we outperformed by $\sim$13.8 and $\sim$11.5 in mIoU and mAcc. metrics, respectively.
While a performance gap remains compared to offline methods~\cite{zhai_cvpr25_panogs,Peng2023OpenScene} that use ground truth point clouds and capture global multi-view information, our method substantially narrows this gap.
Besides, we show some qualitative open-vocabulary 3D semantic segmentation results of ScanNetV2 in~\cref{fig:scannet_semantic_mesh}. 
As shown in the figure, previous online methods only maintained instance-level language features, which led to inaccurate semantic segmentation of open vocabularies due to inaccurate instances and language feature drifts during blending.
Our method can achieve more consistent segmentation results among online approaches.

\myvspace\noindent\textbf{3D Panoptic Segmentation.}
Since some offline baselines (LangSplat~\cite{qin2024langsplat}, OpenGaussian~\cite{wu2024opengaussian}, OpenScene~\cite{Peng2023OpenScene}) marked with $^*$ cannot inherently output 3D instances, PanoGS~\cite{zhai_cvpr25_panogs} provides supervised instance annotations~\cite{vu2022softgroup} for them to compute PRQ metrics (marked in \lo{gray}).
As shown in~\cref{tab:main_results}, compared to the performance gap in 3D semantic segmentation, we further narrow the gap with offline baselines in 3D panoptic segmentation performance, even surpassing PanoGS~\cite{zhai_cvpr25_panogs} on the ScanNetV2~\cite{dai:2017:scannet} dataset. 
The performance also validates the effectiveness of our local-to-global online reconstruction system, which consolidates the 2D inconsistency and obtains robust 3D panoptic map.
More qualitative 3D panoptic segmentation results are provided in our supplementary material.

\myvspace\noindent\textbf{Open Vocabulary Query.}
We show the qualitative open vocabulary query results of OnlineAnySeg~\cite{tang2025onlineanyseg} and our approach in~\cref{fig:exp_open_query}.
The query results of different instances are highlighted with different colors.
While OnlineAnySeg can handle simple queries (\eg, ``television''), it fails on some fine-grained and multi-instance queries (\eg, ``pillow'', ``toilet paper'', ``bag'') due to inaccurate 3D instance segmentation and language feature fusion.
In contrast, our clustering-based approach can obtain accurate instance-level information for objects with the same semantics.
Benefiting from our local-to-global 3D instance modeling approach and fine-grained language feature fusion based on the spatial feature grid, our method can obtain more complete 3D instance information (``bag'') and distinguish different instances with the same semantics (``pillow'') in the scene.

\myvspace\noindent\textbf{Runtime Analysis.}
We evaluate the runtime performance of {\ours} on a desktop computer equipped with an AMD Ryzen 9 7950X CPU and an NVIDIA RTX 4090 GPU.
For \texttt{Scene0645} in the ScanNetV2 dataset, our method takes an average of 410 ms to perform rendering optimization for 5 keyframes with 20 iterations, 350 ms for clustering the 12 keyframes in the sliding window, and 1400 ms for local-to-global matching and fusion (including spatial attribute updates).
Since clustering and fusion process multiple keyframes per sliding window movement (frequency much lower than framerate), our system achieves 18 FPS on simple scenes and 10 FPS on complex scenes (excluding VLM computation for 2D masks and features).
More detailed runtime analyses are provided in our Supp.

\subsection{Ablation Studies}
\label{subsec:ablation}

\begin{figure}[h]
\centering
\includegraphics[width=0.49\linewidth]{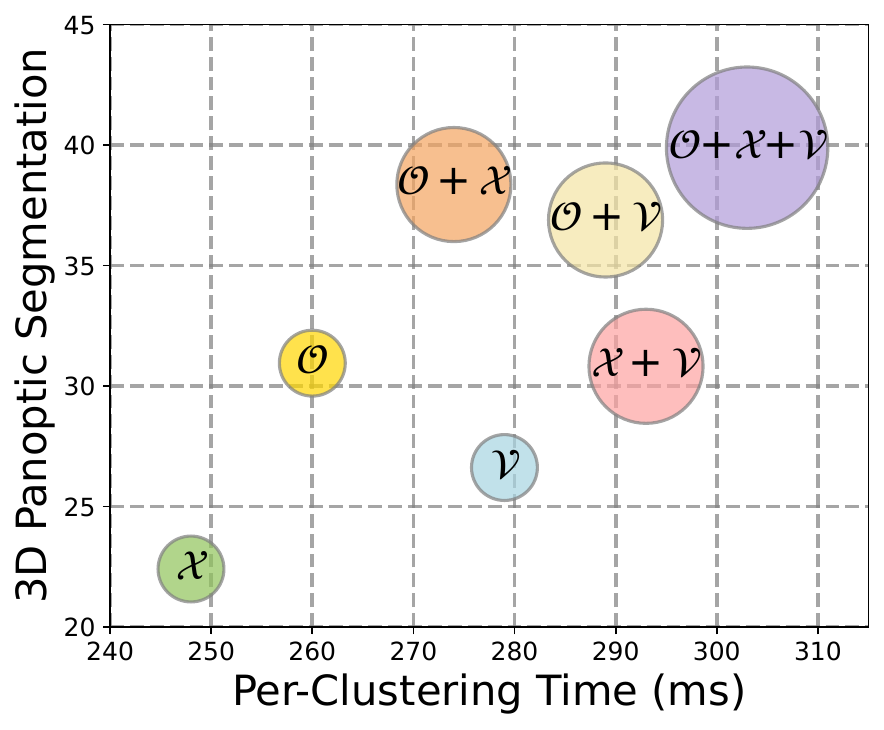}
\includegraphics[width=0.49\linewidth]{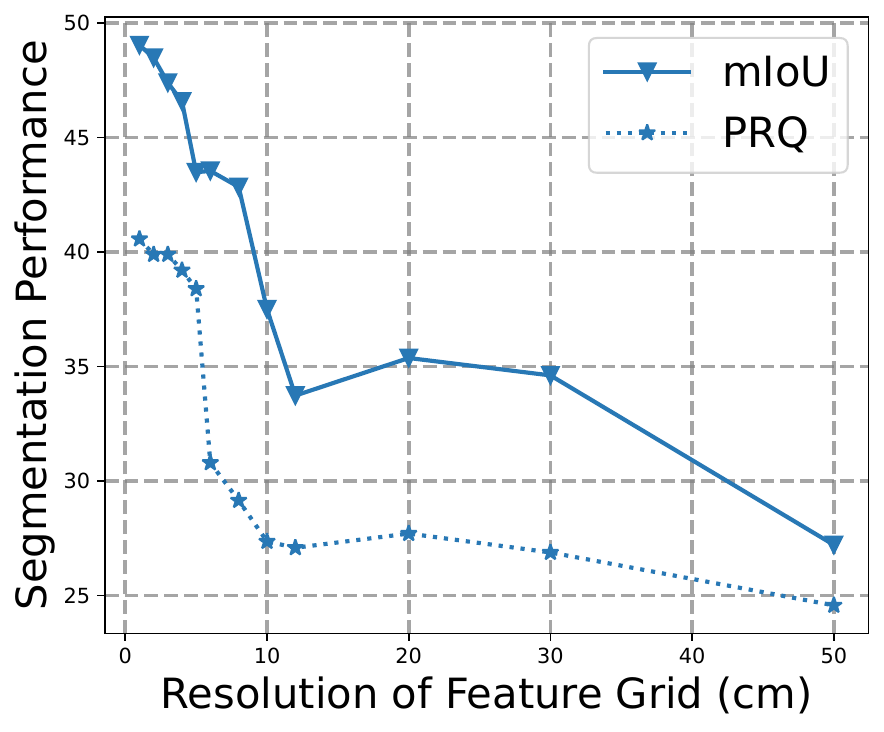}
\vspace{-2mm}
\caption{Left: ablation studies of using different segments clustering cues for local map construction. Right: ablation studies of using different feature grid resolutions. The results are evaluated on ScanNetV2 dataset.}
\label{fig:ablation_cues_resolution}
\end{figure}

\myvspace\noindent\textbf{Effects of Multiple Segments Clustering Cues.} 
To obtain robust and consistent 3D Gaussian instances from sliding windows, we utilized various 3D cues ($\mathcal{O}_{ij}$, $\mathcal{X}_{ij}$, $\mathcal{V}_{ij}$) for graph clustering. 
On the left of~\cref{fig:ablation_cues_resolution}, we show the clustering graph construction time and performance of using different cues during segment clustering to obtain consistent 3D instances. 
As shown in the figure, the horizontal axis represents the average time for clustering operation when using different cues, and the vertical axis represents the 3D panoptic segmentation performance, (PRQ (T) + PRQ (S)) / 2.
Compared to single-cue clustering, multi-cue clustering achieves 8 to 18 PRQ improvement with only $\sim$40 ms additional latency, demonstrating a favorable accuracy-efficiency trade-off.

\begin{table}[h]
\centering
\small
\vspace{-2mm}
\caption{Ablation studies of different matching strategies for global map fusion on ScanNetV2 dataset.}
\vspace{-2mm}
\setlength{\tabcolsep}{0.9em}
\begin{tabular}{lcc}
\toprule
Settings & PRQ (T) & PRQ (S) \\ 
\midrule
\#1 \textit{NN} Match & 24.67 & 22.98 \\
\#2 Only Forward Match $\mathcal {M}_{l \to g}$ & 35.83 & 38.40 \\
\#3 Only Backward Match $\mathcal {M}_{g \to l}$ & 33.71 & 42.72 \\
\#4 Ours Full Bidirectional Match & 37.97        & 41.81 \\
\bottomrule
\end{tabular}
\label{tab:ablation_match}
\end{table}
\myvspace\noindent\textbf{Effects of Bidirectional Bipartite Matching.} 
In~\cref{tab:ablation_match}, we show the performance of different matching strategies for fusing local map from the sliding window into global map.
\#1 represents using the basic nearest neighbor matching algorithm based on the forward matching matrix $\mathcal{M}_{l \to g}$, which obtains the matching relationship between instances in the local map and global map.
Comparing \#1 with others shows that using bidirectional bipartite graph matching can effectively improve accuracy for local-to-global 3D Gaussian instance fusion, with 13.3 and 18.8 improvements for PRQ (T) and PRQ (S) metrics, respectively.
Comparing \#2, \#3 with \#4, we can know that matching results obtained using bidirectional matching are more robust and accurate than those obtained using unidirectional matching.
Furthermore, backward verification particularly benefits \textit{Stuff}-level performance by improving the fusion of newly explored background regions, about 4.3 improvements over forward match for PRQ (S) metric.

\begin{table}[h]
\centering
\small
\vspace{-2mm}
\caption{Ablation studies of our system components. The results are evaluated on ScanNetV2 dataset.}
\vspace{-2mm}
\setlength{\tabcolsep}{0.7em}
\begin{tabular}{lccc}
\toprule
Settings & mIoU & PRQ (T) & PRQ (S) \\ 
\midrule
\#1 \textit{w/o.} Segment Clustering & 48.48  & 32.25 & 30.68  \\ 
\#2 \textit{w/o.} Feature Grid $\mathcal{F}$ & 30.40 & 26.71 & 24.92 \\ 
\#3 Ours Full & 48.48 & 37.97 & 41.81  \\
\bottomrule
\end{tabular}
\label{tab:ablation_l2g}
\end{table}

\begin{figure}[h]
\centering
\includegraphics[width=\linewidth]{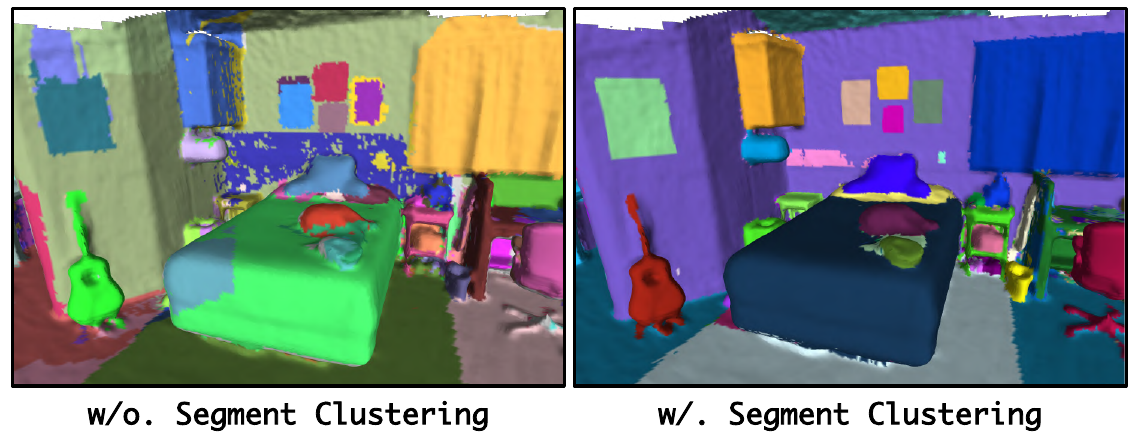}
\vspace{-7mm}
\caption{Qualitative comparison of using segment clustering.}
\vspace{-2mm}
\label{fig:ablation_clustering}
\end{figure}

\myvspace\noindent\textbf{Effects of System Components.} 
In~\cref{tab:ablation_l2g}, we analyze the effectiveness of our segment clustering and feature grid $\mathcal{F}$ inside the spatial attribute.
\textit{1) About Segment Clustering:} \#1 indicates that we directly fuse keyframe segments into the global map without clustering.
Comparing \#1 and \#3, we can see that directly fusing noisy segments leads to incorrect segmentation of 3D instances, leading to poor panoptic understanding, especially for large \textit{Stuff} regions.
Besides, the qualitative 3D instance results of using segment clustering from \texttt{Scene0000} are shown in~\cref{fig:ablation_clustering}.
Without performing segment clustering over the local sliding window $\mathcal{W}$, 3D instances in the global map will contain a lot of noise.
\textit{2) About Feature Grid:} \#2 indicates we maintain a coarse language feature for each 3D instance without using feature grid $\mathcal{F}$. 
Comparing \#2 and \#3, we can know that our spatial attribute module can improve the open-vocabulary scene understanding performance of our system.
Besides, we investigate the performance of the spatial resolution used for feature grids $\mathcal{F}$.
The results on the ScanNetV2 dataset are shown in the right of~\cref{fig:ablation_cues_resolution}, which show that spatial resolution significantly impacts performance: resolutions coarser than 5 cm cause rapid performance degradation and convergence to a lower bound.

\section{Conclusion}
\label{sec:conclusion}
We present {\ours}, a novel and effective system for online open-vocabulary panoptic mapping using 3D Gaussian Splatting.
Our approach employs a local-to-global paradigm with sliding windows to achieve efficient online reconstruction and perception.
Key contributions include: (1) a multi-cue segment clustering algorithm that fuses noisy 2D priors into consistent 3D instances, (2) explicit spatial attribute grids for semantic feature storage, and (3) bidirectional bipartite matching for robust local-to-global fusion.
Extensive experiments demonstrate that {\ours} outperforms existing online methods for open-vocabulary 3D panoptic scene understanding.

\myvspace\noindent\textbf{Limitations:} 
(1) Our method currently cannot reconstruct dynamic objects.
(2) Like previous methods~\cite{zhai_cvpr25_panogs,tang2025onlineanyseg,o2v-mapping_eccv24}, we require depth and pose inputs.
Our future work will explore feed-forward approaches~\cite{wang2024dust3r,wang2025vggt,liu2025slam3r} that eliminate these requirements for fully pose-free and depth-free open-vocabulary reconstruction.

\noindent\textbf{Acknowledgment:} This work was partially supported by NSF of China (No. 62425209).


{
    \small
    \bibliographystyle{ieeenat_fullname}
    \bibliography{draft}
}

\clearpage
\setcounter{page}{1}

\twocolumn[{
    \renewcommand\twocolumn[1][]{#1}%
    \maketitlesupplementary
    \centering
    \includegraphics[width=\linewidth]{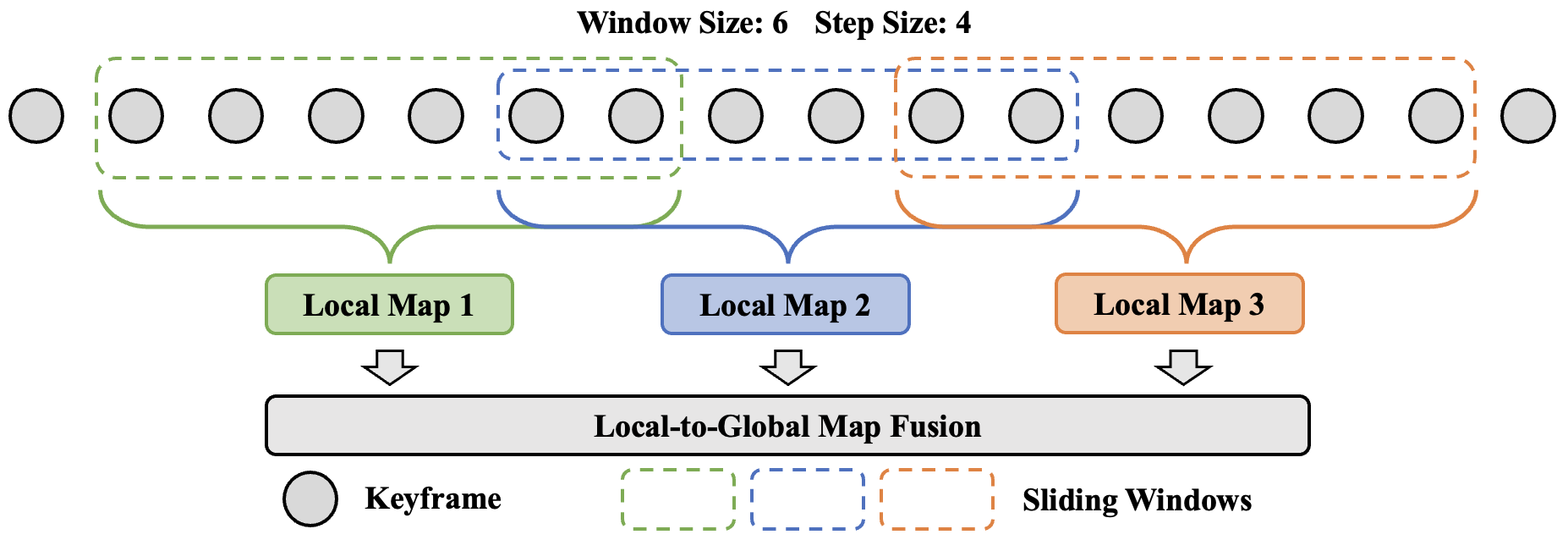}
\captionof{figure}{\textbf{Illustration of Sliding Window}. For a sliding window with a size of 6 and a step size of 4, we perform segment clustering every 4 keyframes and merge the local map into the global map. As shown in the figure, this reduces the reading and updating of the global map, enabling efficient reconstruction.}
\label{fig:supp_sliding_window}    
\vspace{4mm}
}]

In this supplementary document, we first provide more implementation details in~\cref{sec:implement}. 
Next, we supply more visualization results of our methods in~\cref{sec:more_results}.

\section{Implementation Details}
\label{sec:implement}

\myvspace\noindent\textbf{Dataset Setting} 
For ScanNetV2~\cite{dai:2017:scannet}, we use the following 10 scenes:
\textit{scene0000}, \textit{scene0062}, \textit{scene0070}, \textit{scene0097}, \textit{scene0140}, \textit{scene0200}, \textit{scene0347}, \textit{scene0400}, \textit{scene0590}, and \textit{scene0645}. 
All selected scenes are evaluated on the \textit{00} trajectory.
For the evaluation of 3D semantic and panoptic segmentation, we use the 19 classes: 
\textit{wall}, \textit{floor}, \textit{cabinet}, \textit{bed}, \textit{sofa}, \textit{table}, \textit{door}, \textit{window}, \textit{bookshelf}, \textit{picture}, \textit{counter}, \textit{desk}, \textit{curtain}, \textit{refrigerator}, \textit{shower curtain}, \textit{toilet}, \textit{sink}, and \textit{bathtub}.
For Replica dataset~\cite{julian:2019:replica}, the commonly-used 8 scenes $\{\textit{room0-2},\textit{office0-4}\}$ are used for evaluation, and two additional labels, \textit{other furniture} and \textit{ceiling}, are used for evaluation.

\begin{figure*}[ht!]
\centering
\includegraphics[width=\linewidth]{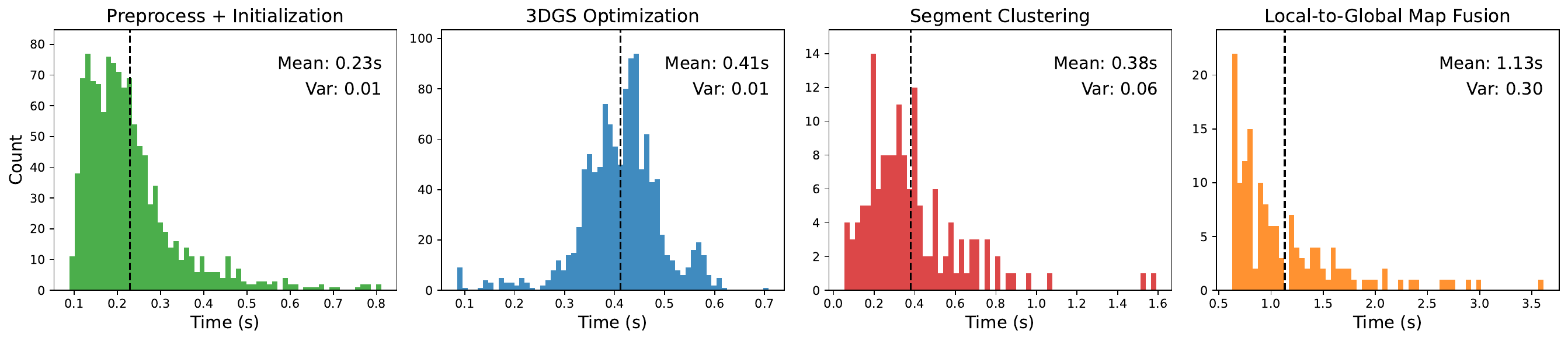} 
\caption{\textbf{Runtime Performance of OnlinePG}. We divided the system's time consumption into four parts and statistically analyzed the distribution of the execution time of each part across ten scenarios used by ScanNet. The black dashed line represents the average time.}
\label{fig:supp_time_component}
\end{figure*}

\myvspace\noindent\textbf{Details of Our OnlinePG}
Following~\cite{Peng2023OpenScene,zhai_cvpr25_panogs,wu2024opengaussian}, we adopt CLIP~\cite{clip} and LSeg~\cite{Lseg} as text and image visual-language feature extractors, with feature dimension $D_f=512$.
We use EntitySeg~\cite{cropformer} to extract 2D instance segmentation for each keyframe.
We sample a keyframe every 20 frames and maintain a sliding window of size 12.
Segment clustering and local-to-global map fusion are performed every 7 keyframes.
The illustration of sliding window design is shown in~\cref{fig:supp_sliding_window}.
For an online image stream, we maintain a fixed-size window and move it in fixed steps. 
Each time the sliding window moves, it builds a local map for all keyframes in the window and merges it into the global map based on bidirectional matching.
The resolutions of the Replica~\cite{julian:2019:replica} and ScanNetV2~\cite{dai:2017:scannet} datasets that we used are 640$\times$360 and 640$\times$480, respectively.
When performing bidirectional bipartite matching between the local and global maps, we remove matches with scores below the threshold $1/N_{ins}$ to filter out erroneous results, where $N_{ins}$ is the number of candidate matches.
For rendering optimization, the learning rates for the 3D Gaussian's location, opacity, scale, color, and rotation are set to 0.00015, 0.05, 0.001, 0.001, 0.01, respectively.

\myvspace\noindent\textbf{Runtime Analysis} 
Because the running speed of our method is affected by the number of masks in the 2D image and the number of instances in the 3D scene, we report the running time in 10 selected scenes from ScanNetV2.
For the video stream, we select one frame every 20 frames as a keyframe and process it.
As shown in~\cref{fig:supp_time_component}, we show the runtime of four main parts in our system: {(\#1) Keyframe Preprocessing and 3D Segments Initialization}, {(\#2) 3DGS Optimization}, {(\#3) Segment Clustering}, and {(\#4) Local-to-Global Map Fusion}.
\#1 and \#2 are called once every time a new keyframe is inserted.
\#3 and \#4 are called only when the sliding window moves a fixed step size.

Besides, the FPS performance of different online methods is also shown in~\cref{tab:supp_replica_performance} and~\cref{tab:supp_scannet_performance}.
Since OnlineAnySeg~\cite{tang2025onlineanyseg} needs to obtain the mask and features in advance, the FPS results reported by different methods in the table do not include the time of VLM inference.

\section{More Experimental Results}
\label{sec:more_results}

\myvspace\noindent\textbf{Detailed Performance of Each Scene}
The detailed 3D semantic and panoptic segmentation performance of our approach and different online baselines (O2V-Mapping~\cite{o2v-mapping_eccv24} and OnlineAnySeg~\cite{tang2025onlineanyseg}) are shown in~\cref{tab:supp_replica_performance} and~\cref{tab:supp_scannet_performance}, respectively.

\begin{figure}[ht!]
\centering
\includegraphics[width=\linewidth]{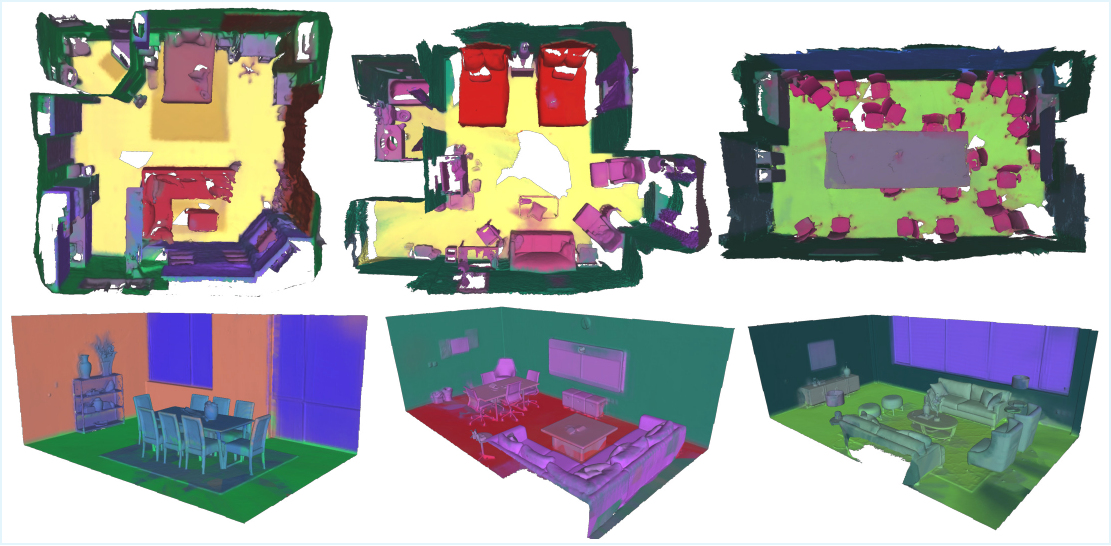} 
\caption{\textbf{Visualization Results of 3D Language Features}. For better visualization, we perform principal components analysis (PCA) on the high-dimensional language features.}
\label{fig:supp_feature}
\end{figure}
\myvspace\noindent\textbf{3D Language Feature}
In~\cref{fig:supp_feature}, we show the visualization results of the language features reconstructed by our OnlinePG. 
We used principal component analysis to compress the 512-dimensional features into 3 dimensions for visualization. 
As can be seen from the figure, the features we obtained can distinguish finer-grained objects, such as carpet and floor, bed and sofa, \etc.
Objects with semantic similarity have similar characteristics, and visualization can show that they have similar colors.

\begin{figure}[ht!]
\centering
\includegraphics[width=\linewidth]{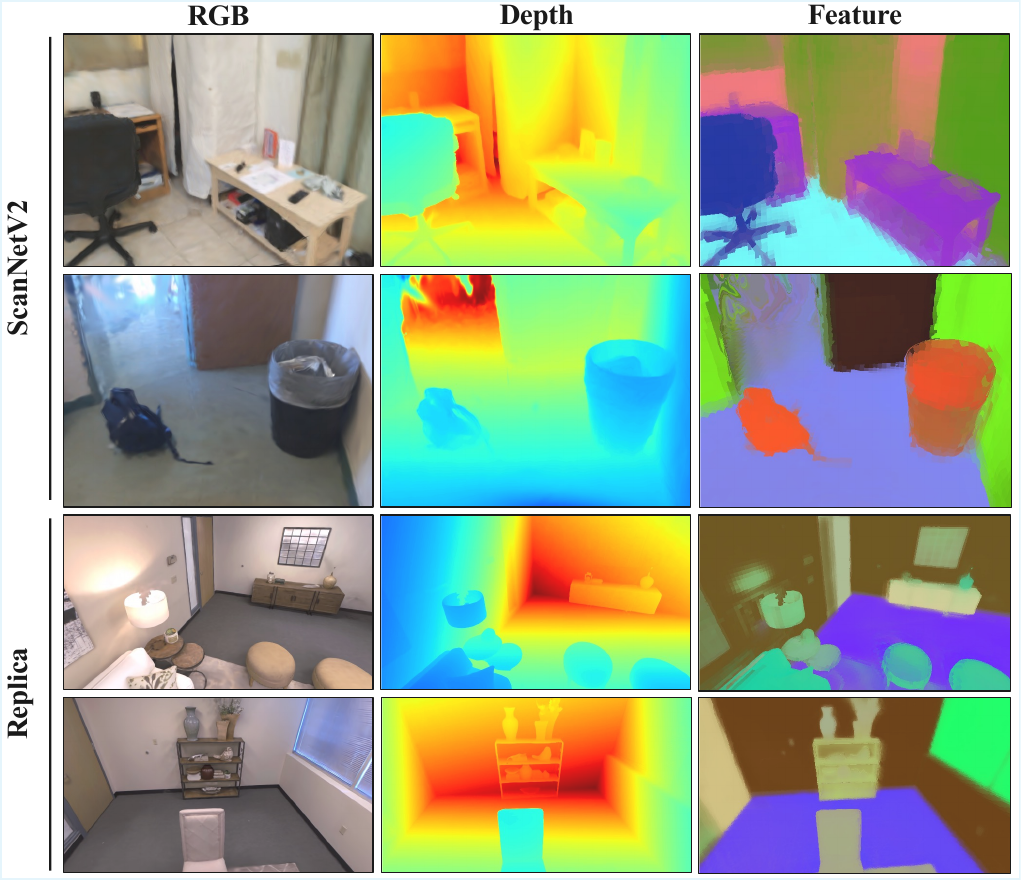} 
\caption{\textbf{Qualitative Rendering Results}. We show the rendered RGB, Depth, and language feature of different scenes.}
\label{fig:supp_rendering}
\end{figure}
\myvspace\noindent\textbf{Rendering Results}
In~\cref{fig:supp_rendering}, we present rendering results from four viewpoints on the ScanNetV2~\cite{dai:2017:scannet} and Replica~\cite{julian:2019:replica} datasets, including RGB, depth, and language features.

\begin{figure*}[h]
\centering
\includegraphics[width=\linewidth]{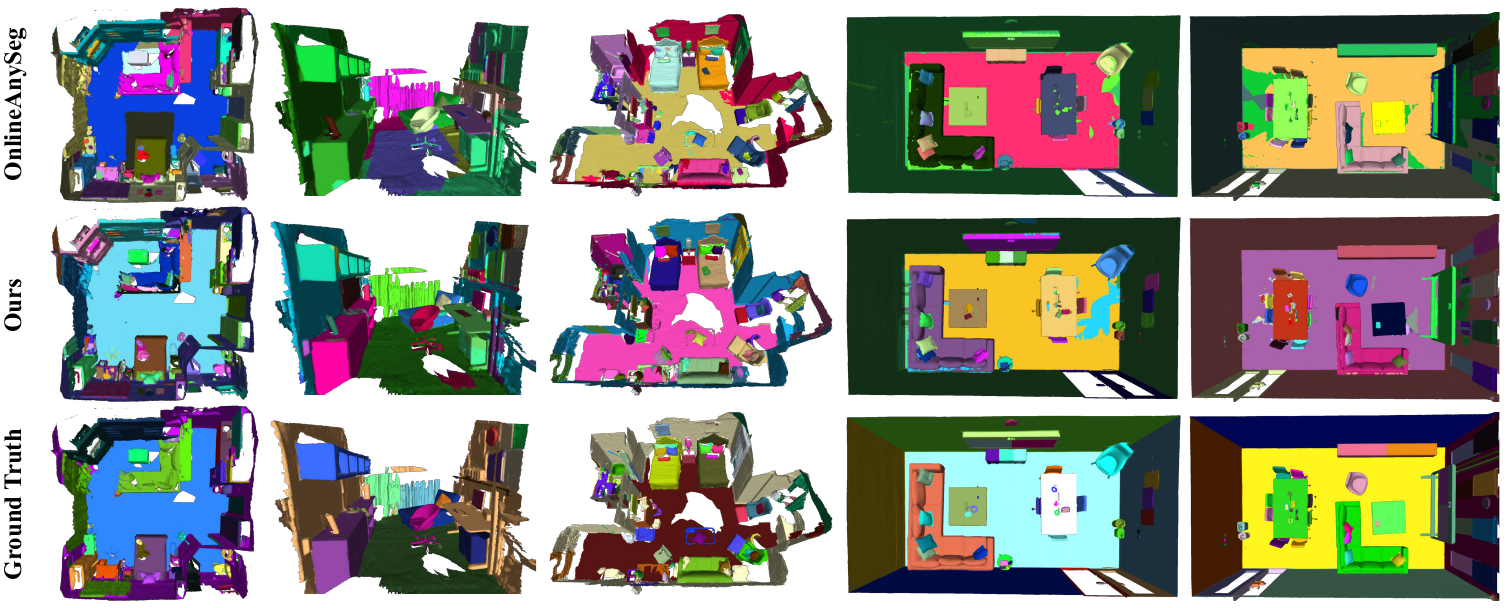} 
\vspace{-4mm}
\caption{\textbf{Qualitative 3D Instance Segmentation Comparison}. We show the 3D instance segmentation results from several scenes from ScanNetV2~\cite{dai:2017:scannet} and Replica~\cite{julian:2019:replica} datasets.}
\label{fig:supp_3d_instance}
\end{figure*}

\myvspace\noindent\textbf{3D Instance Results}
In~\cref{fig:supp_3d_instance}, we present 3D instance segmentation results of some scene from the ScanNetV2~\cite{dai:2017:scannet} and Replica~\cite{julian:2019:replica} datasets. 
Different colors represent different 3D instances.
Due to the GT mesh of the Replica dataset containing many unobserved areas in the images, this will lead to some noisy instances (on the floor and walls), which will also cause the PRQ (S) of the online method to be worse than that of the offline method.

\begin{table*}[!tp]
\renewcommand\arraystretch{0.9}
\centering
\caption{\textbf{3D Semantic and Panoptic Segmentation Results on Replica Datasets}. We demonstrate the 3D segmentation performance and FPS data of the online method in different scenes of Replica~\cite{julian:2019:replica}.}
\vspace{-2mm}
\small
\setlength{\tabcolsep}{8.33pt}
\begin{tabularx}{\linewidth}{llcccccccc}
\toprule
Method & Metrics & Room 0 & Room 1 & Room 2 & Office 0 & Office 1 & Office 2 & Office 3 & Office 4 \\
\midrule
\multirow{3}{*}{O2V-Mapping~\cite{o2v-mapping_eccv24}} 
& mIoU    & 29.25 & 39.58 & 32.53 & 18.78 & 17.54 & 20.75 & 12.59 & 23.24 \\
& mAcc.   & 38.00 & 52.60 & 43.22 & 27.56 & 20.25 & 30.20 & 17.95 & 38.47 \\
& FPS &  3.08 & 3.05 & 3.08 & 3.05 & 3.10 & 3.13 & 3.18 & 3.28 \\
\midrule
\multirow{5}{*}{OnlineAnySeg~\cite{tang2025onlineanyseg}} 
& mIoU & 46.76 & 45.13 & 51.30 & 37.54 & 25.38 & 32.91 & 25.45 & 48.23 \\
& mAcc. & 62.41 & 67.50 & 66.84 & 50.47 & 31.24 & 46.99 & 42.46 & 59.45 \\
& PRQ (T) & 30.26 & 51.65 & 26.59 & 45.34 & 22.10 & 24.90 & 16.32 & 27.17 \\
& PRQ (S) & 8.58 & 15.99 & 10.97 & 8.65 & 2.73 & 7.15 & 4.83 & 11.46 \\
& FPS     & 9.90 & 13.06 & 12.91 & 15.40 & 20.94 & 11.65 & 10.81 & 11.45 \\
\midrule
\multirow{5}{*}{Ours} 
& mIoU & 53.60 & 67.16 & 68.31 & 32.07 & 10.19 & 53.53 & 54.95 & 43.55 \\
& mAcc. & 59.34 & 80.17 & 75.86 & 36.28 & 16.07 & 60.06 & 60.86 & 50.96 \\
& PRQ (T) & 31.03 & 57.49 & 51.14 & 48.45 & 0.00 & 48.40 & 47.46 & 43.89 \\
& PRQ (S) & 11.75 & 17.22 & 12.44 & 7.37 & 4.47 & 17.48 & 9.20 & 18.05 \\
& FPS     & 12.07 & 13.91 & 14.07 & 15.85 & 18.22 & 14.77 & 12.42 & 14.14 \\
\bottomrule
\end{tabularx}
\label{tab:supp_replica_performance}
\end{table*}
\begin{table*}[!tp]
\centering
\small
\renewcommand\arraystretch{0.9}
\setlength{\tabcolsep}{8pt}
\caption{\textbf{3D Semantic and Panoptic Segmentation Results on ScanNetV2 Datasets}. We demonstrate the 3D segmentation performance and FPS data of the online method in different scenes of ScanNetV2~\cite{dai:2017:scannet}.}
\vspace{-2mm}
\begin{tabularx}{\linewidth}{llcccccccccc}
\toprule
Method & Metrics & 0000 & 0062 & 0070 & 0097 & 0140 & 0200 & 0347 & 0400 & 0590 & 0645 \\
\midrule
\multirow{3}{*}{O2V-Mapping~\cite{o2v-mapping_eccv24}} 
& mIoU    & 35.80 & 46.09 & 41.21 & 26.29 & 24.25 & 39.08 & 27.54 & 39.74 & 33.18 & 28.27 \\
& mAcc.   & 59.92 & 72.72 & 55.59 & 55.73 & 34.15 & 63.91 & 47.12 & 58.33 & 49.85 & 52.22 \\
& FPS & 3.38 & 3.43 & 3.43 & 3.53 & 3.40 & 3.50 & 3.43 & 3.38 & 3.43 & 3.38 \\
\midrule
\multirow{5}{*}{OnlineAnySeg~\cite{tang2025onlineanyseg}} 
& mIoU    & 32.64 & 31.13 & 23.26 & 39.08 & 12.62 & 39.32 & 32.97 & 33.93 & 34.66 & 33.26 \\
& mAcc.   & 50.61 & 60.73 & 31.92 & 69.62 & 31.86 & 52.43 & 54.51 & 56.42 & 55.89 & 58.02 \\
& PRQ (T) & 42.32 & 33.47 & 36.29 & 71.30 & 53.32 & 33.76 & 38.43 & 25.97 & 38.95 & 56.06 \\
& PRQ (S) & 18.53 & 38.20 & 12.89 & 41.70 & 7.94  & 39.09 & 20.35 & 38.26 & 24.50 & 21.32 \\
& FPS     & 15.57 & 22.91 & 26.44 & 22.83 & 15.72 & 19.27 & 20.41 & 25.22 & 22.98 & 16.93 \\
\midrule
\multirow{5}{*}{Ours} 
& mIoU    & 39.75 & 75.09 & 46.50 & 52.59 & 48.04 & 47.84 & 52.76 & 39.44 & 37.95 & 44.92 \\
& mAcc.   & 61.77 & 90.36 & 62.95 & 70.39 & 64.54 & 62.11 & 72.50 & 62.18 & 53.33 & 60.05 \\
& PRQ (T) & 20.21 & 35.28 & 49.55 & 58.63 & 51.67 & 46.56 & 42.08 & 0.00 & 33.55 & 42.16 \\
& PRQ (S) & 40.05 & 63.27 & 24.34 & 52.79 & 39.28 & 43.94 & 36.68 & 56.53 & 33.36 & 27.88 \\
& FPS     & 13.33 & 18.83 & 15.68 & 17.57 & 14.23 & 18.48 & 17.86 & 16.89 & 15.26 & 14.08  \\
\bottomrule
\end{tabularx}
\label{tab:supp_scannet_performance}
\end{table*}

\end{document}